\documentclass[a4paper]{article}

\usepackage{INTERSPEECH2021}
\usepackage{enumitem}
\usepackage{hyperref}
\usepackage{soul}
\usepackage{listings}
\usepackage{cite}

\title{Integrating Dialog History into End-to-End Spoken \\Language Understanding Systems}
\name{Jatin Ganhotra\textsuperscript{*}, Samuel Thomas\textsuperscript{*}, Hong-Kwang J. Kuo, Sachindra Joshi, \\George Saon, Zolt{\'a}n T\"{u}ske, Brian Kingsbury}
\address{
 IBM Research AI, Yorktown Heights, NY, USA
 }
\email{}

\begin{document}

\maketitle

\begin{NoHyper}
\begingroup\renewcommand\thefootnote{*}
\footnotetext{Equal contribution}
\endgroup
\end{NoHyper}

\begin{abstract}
End-to-end spoken language understanding (SLU) systems that process human-human or human-computer interactions are often context independent and process each turn of a conversation independently. Spoken conversations on the other hand, are very much context dependent, and dialog history contains useful information that can improve the processing of each conversational turn. In this paper, we investigate the importance of dialog history and how it can be effectively integrated into end-to-end SLU systems. While processing a spoken utterance, our proposed RNN transducer (RNN-T) based SLU model has access to its dialog history in the form of decoded transcripts and SLU labels of previous turns. We encode the dialog history as BERT embeddings, and use them as an additional input to the SLU model along with the speech features for the current utterance.
We evaluate our approach on a recently released spoken dialog data set, the \textsc{HarperValleyBank} corpus. We observe significant improvements: 8\% for dialog action and 30\% for caller intent recognition tasks, in comparison to a competitive context independent end-to-end baseline system.
\end{abstract}

\section{Introduction}
Conventional spoken language understanding (SLU) systems have been built by integrating a text-based natural language understanding system with an automatic speech recognition (ASR) system~\cite{Goel2005,Haghani2018}.
Recently, these traditional systems have been replaced by end-to-end (E2E) systems that directly process speech to produce spoken language understanding (SLU) entity or intent label targets without any intermediate ASR processing~\cite{serdyuk2018towards,qian2017exploring,chen2018spoken,lugosch2019speech,caubriere2019curriculum,huang2020leveraging,price2020improved,radfar2020end,tian2020improving,jia2020large,kuo2020end,palogiannidi2020end}. When processing human-human or human-computer interactions, these E2E SLU systems process each turn of a conversation independently. However, spoken task-oriented conversations are often context dependent as users and agents converse in multi-turn conversations to achieve the various user goals. These turns are  also related, as the user or agent might refer to information introduced in previous turns. Without proper context these pieces of information  introduce ambiguity. For example, “one” could refer to a scheduled appointment date or a part of a phone number or zip code depending on the context. Dialog history hence contains useful information that can be effectively used to improve the processing of each conversational turn and resolve such ambiguities \cite{bhargava2013easy, xu2014contextual}. 

Several previous approaches have studied the importance of dialog history and how it can be effectively encoded for use with conventional text-based SLU systems \cite{chen2016end,sankar2019neural,goel2019hyst,vukotic2016step}. A general theme of these works has been to use embeddings to represent utterances or dialog turns preceding the current utterance that is being processed \cite{henaff2016tracking,korpusik2019dialogue}. Some prior work has explored encoding the full dialog history using Recurrent Neural Network (RNN) \cite{gupta2018efficient} and Graph Convolutional Network based approaches \cite{9330801} but all these approaches have primarily  been in the context of a conventional text-based SLU system. 

\begin{figure}
    \centering
    \includegraphics[scale=0.39]{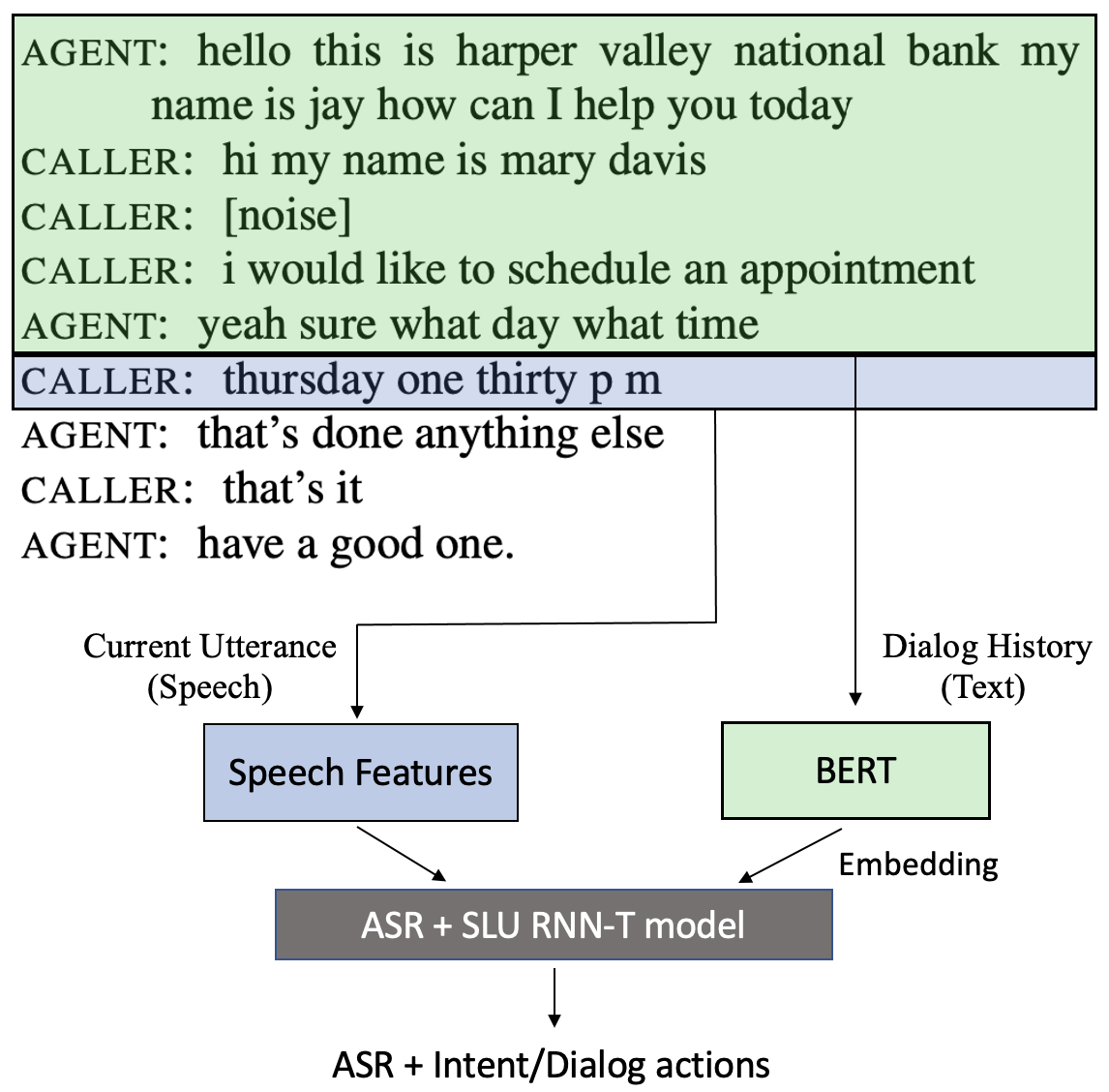}
    \caption{Our proposed E2E SLU model with dialog history}
    \label{fig:arch}
\end{figure}

Recently, in \cite{tomashenko2020dialogue} the benefits of integrating dialog history into a speech based E2E SLU system have been investigated. The authors explore different representations, both supervised and unsupervised, to generate dialog history embedding vectors, and they observe significant improvements on the semantic slot filling task. However, this initial attempt in integrating dialog history into an E2E SLU system is limited, as the authors only use the most recent system prompt before the current utterance and do not consider the full dialog history.

In this paper, we revisit the problem of how to effectively encode full dialog history into a speech based E2E SLU system. We use the \textsc{HarperValleyBank} corpus \cite{wu2020harpervalleybank} for our experiments. The dataset is a free, public domain spoken dialog corpus, where the dialogs simulate simple consumer banking interactions between users and agents. Using this corpus, we study the use of BERT \cite{devlin2019bert} embeddings to encode various elements of dialog history: the textual content of previous turns, speaker role (whether agent or user) for each turn and previous SLU tags for each utterance in the dialog history. These embeddings are then used as features that contain side information on dialog history for an RNN Transducer based E2E SLU system\cite{sam2021rnn,george2021rnn} as shown in Fig.\ref{fig:arch}. This use of side information is similar to past approaches where speaker identity has been provided as i-vector embeddings \cite{saon2013speaker} for ASR systems or h-vector embeddings for dialog history in SLU systems \cite{tomashenko2020dialogue}. We conduct experiments on two SLU tasks, dialog action prediction and intent recognition, to demonstrate the usefulness of our proposed method in integrating dialog history for E2E SLU systems.


\section{Approach}
\label{sec:dialog-history-and-slu}

\subsection{RNN-T based SLU systems}
RNN-T based ASR models are a class of single end-to-end trained, streamable, all-neural models that have been widely adopted for speech recognition~\cite{he2019streaming,rao2017exploring,li2019improving,shafey2019joint,ghodsi2020rnn}. These models typically consist of three different sub-networks: a transcription network, a prediction network, and a joint network~\cite{graves2012sequence}. The transcription network produces acoustic embeddings, while the prediction network resembles a language model in that it is conditioned on previous non-blank symbols produced by the model. The joint network combines the two embedding outputs to produce a posterior distribution over the output symbols. This architecture elegantly replaces a conventional ASR system composed of separate acoustic model, language model, pronunciation lexicon, and decoder components.

Given their popularity, and the fact that RNN-T models can naturally handle more abstract output symbols such as ones marking speaker turns~\cite{shafey2019joint}, in our previous work \cite{sam2021rnn} we developed extensions of these models for SLU tasks. The RNN-T models we develop for SLU are first pre-trained on task independent ASR data. The pre-trained model is then modified to include semantic labels specific to the SLU task by resizing the output layer and the embedding layer of the prediction network to include additional output nodes. 

\subsection{Encoding Dialog History} 



Recently, BERT-based models have shown great promise for multi-turn dialog tasks \cite{whang2020effective, gu2020speaker}. BERT is a bidirectional model based on the transformer architecture \cite{devlin2019bert} and is pre-trained on two unsupervised tasks: masked language modeling (MLM) and next sentence prediction (NSP). The pre-trained BERT model can be fine-tuned later on a downstream task, defined as follows. Given a dialog dataset $\mathcal{D}$, an example is denoted as a triplet \textless $c,u_t,l$ \textgreater, where $c = \{u_1,u_2,...,u_{t-1}\}$ represents the dialog context with $t-1$ utterances ($u$), $u_t$ is the current utterance, and $l$ denotes a label. $l$ supplies the correct intent for an intent recognition task or the set of correct dialog actions for a dialog actions prediction task.

We use BERT as our base model for encoding dialog history and explore the impact of additional information from the context on the downstream task. We use the uncased BERT-base model\footnote{\url{https://huggingface.co/bert-base-uncased}} for our experiments and \texttt{BertForSequenceClassification} operating as a binary classifier as follows. The dialog context $c$ and current utterance $u_t$ are fed together to BERT as a sequence ([CLS] $c$ [SEP] $u_t$ [SEP]) for fine-tuning the model. We explore different strategies for encoding additional information from $c$, such as speaker role identifiers and dialog acts for previous utterances, and investigate their impact on the downstream task. After fine-tuning, for each utterance $u_t$ at turn $t$, we use the sequence ([CLS] $c$ [SEP]) as input to the model and use the hidden state of the [CLS] token $c_{t}^{CLS}$ from the model as the embedding representation of the dialog context $c$ at turn $t$ for the SLU model. 






\subsection{Integrating Dialog History}
To integrate the BERT based embedding described above, in addition to modifying the output layers of the network, the input layers of the network are also adapted. While baseline SLU models are trained on conventional log-mel acoustic features, our proposed models that integrate dialog history have BERT based embeddings appended to their input acoustic features as well. To accommodate the increase in the input dimension, the input layer of the transcription network that receives the input features is expanded to include the embedding feature dimensions. The new network parameters are randomly initialized, while the remaining parts are initialized from the pre-trained ASR network. After these modifications, the proposed model is trained on SLU data similar to a regular ASR based model \cite{george2021rnn}.

\section{Experiments and Results}

\subsection{Dataset}
We use the \textsc{HarperValleyBank} corpus for our experiments \cite{wu2020harpervalleybank}. 
The dataset is a free, public domain spoken dialog corpus, where the dialogs simulate simple consumer banking interactions between users and agents. There are 1,446 human-human conversations (23 hours of audio) between 59 unique speakers. There are transcripts provided for each utterance, in addition to annotations for speaker identity, caller intent, dialog actions, and
emotional valence.

We focus on two tasks in our work: caller intent and dialog action prediction. 
The caller intent task attempts to predict a single intent that represents the customer’s goal in the conversation. Each conversation is labelled with one of eight categories: order checks, check balance, replace card, reset password, get branch hours, pay bill, schedule appointment, or transfer money \cite{wu2020harpervalleybank}. While the caller intent is an SLU annotation at the conversation level, each utterance in a conversation is labelled with one or more dialog actions. The set of all 16 possible dialog actions is: 
“yes” response, greeting, response, data confirmation, procedure explanation, data question, closing, data communication, “bear with me” response, acknowledgement, data response, filler disfluency, thanks, open question, problem description, and other \cite{wu2020harpervalleybank}. 

Before the various BERT and SLU models are constructed, transcripts of the original data are processed to remove non-lexical tokens such as [noise], [laughter] and $<$unk$>$. The processed transcripts are then aligned with the audio and significant non-speech regions are marked out. The final data set we use for our experiments corresponds to about 12.5 hours of speech. This data set is partitioned into an ASR train, validation and test set based on the same speaker ids used for ASR experiments in \cite{wu2020harpervalleybank}. The ASR train set contains 10.3 hours of data (48 unique speakers), the validation set contains 0.7 hours of data (5 unique speakers) and the test set contains 1.5 hours of data with 6 unique speakers. For our dialog history based SLU experiments, instead of creating a completely new test set, we expand the ASR test set into full conversations with both agents and callers. Note that in the corpus, a person could have participated as an agent or a caller in different calls. We construct the SLU test set so that the callers are all unseen speakers while the agents are part of the training set. The SLU test set contains 1.8 hours of data, corresponding to 199 full conversations between 48 speakers. To better train the proposed E2E models, additional copies of the corpus are created using speed/tempo perturbation~\cite{Ko15}. The ASR training corpus after data augmentation is about 115 hours of data.

\subsection{BERT based embeddings for SLU} 
\label{bert-training-details}

We modify the original BERT architecture by adding a linear layer before the binary classifier (\texttt{BertForSequenceClassification}), to reduce the [CLS] embedding dimension to 128. This reduced dimensionality is empirically chosen to achieve a balance between the 240 dimensional input speech features and the dialog context embedding.  The dimensionality reduction does, however, result in a performance drop of about 2\% for the BERT classifiers. We train separate models for the intent and dialog action classification tasks. These models are trained for 20 epochs using an AdamW optimizer, a batch size of 32 and learning rate set to 2e-5. We use sigmoid cross entropy loss for dialogue act classification, since it is a  multi-label binary classification task, and softmax cross entropy loss for intent classification.

Baseline models for our experiments are trained in the default setting using only the current utterance without any previous dialog history. The results for these models are reported as [A1] and [B1] in Table \ref{text-results-intent-table-sam-dataset} and Table \ref{text-results-dialog-act-table-sam-dataset} respectively. 
In addition to the baseline models, we investigate the impact of dialog history using different strategies to encode additional information, such as  speaker role identifiers and dialog actions for previous utterances. For a new turn $t$, the additional input features from the dialog context $c$, where $c = \{u_1,u_2,...,u_{t-1}\}$ are:

\begin{itemize}
    \item $[$SPEAKER$]$ - Additional speaker tokens $<$user$>$ and $<$agent$>$ are used to identify the speaker for each utterance $u$ in the dialog history . 
    \item $[$HISTORY$]$ - The original text of all previous utterances in the dialog history is used, i.e. $\{u_1,u_2,...,u_{t-1}\}$
    \item $[$DIALOG-ACTS$]$ - The ground-truth dialog actions for utterances in the dialog history are used for the dialog action classification task, i.e. $\{da_1,da_2,...,da_{t-1}\}$
\end{itemize}

\noindent The additional speaker tokens ($<$user$>$, $<$agent$>$) and the 16 unique tokens for dialog actions are added as special tokens to the BERT Tokenizer vocabulary during training. For each utterance in the data set, a unique history is constructed in three different ways by using the current utterance text and additional information from dialog history:

\begin{enumerate}
    \item $[$SPEAKER$]$ + $[$HISTORY$]$ - to evaluate the impact of original text from dialog history. 
    \item $[$SPEAKER$]$ + $[$DIALOG-ACTS$]$ - to evaluate the impact of dialog acts and validate if the dialog acts are a good substitute for the original text for an utterance.
    \item $[$SPEAKER$]$ + $[$HISTORY$]$ + $[$DIALOG-ACTS$]$ - to evaluate the performance when all available additional information is used.
\end{enumerate}

\newcommand*{\MyIndent}{\hspace*{0.2cm}}
\newcommand*{\MyIndentA}{\hspace*{0.5cm}}


\begin{table}[t]
\caption{Evaluation results of BERT for intent classification.}



\centering
\begin{tabular}{cl|c}
\toprule
\multicolumn{2}{c}{Input Features} & Accuracy (\%) \\
\hline \hline
[A1] & Current-utterance & 49.0 \\ \hline
[A2] & + $[$SPEAKER$]$ + $[$HISTORY$]$ & 86.6\\ \hline
\end{tabular}
\label{text-results-intent-table-sam-dataset}
\vspace{-10pt}
\end{table}

From Table \ref{text-results-intent-table-sam-dataset}, we observe that using $[$SPEAKER$]$ + $[$HISTORY$]$ in creating dialog history embeddings leads to a significant improvement in performance [A2]  compared to the baseline [A1] which uses just the current utterance. A similar improved performance on the dialog act classification task [B2] in comparison to the baseline [B1] can be seen in Table \ref{text-results-dialog-act-table-sam-dataset} with this configuration. We notice that using dialog-acts as a substitute for the original text utterances does improve performance [B3], but is lower than [B2]. Finally, using both original text and the corresponding dialog acts achieves the best performance for the dialog-act classification task [B4]. We further explore the span of dialog history needed for these embeddings in two additional experiments [B5] and [B6], where we only consider the previous 4 or 2 utterances instead of the full dialog history. We observe that the last few utterances do not capture enough additional information to be used as a substitute for the entire dialog history for the dialog-act classification task.

\subsection{RNN-T based SLU systems with Dialog History} 

As described earlier, the RNN-T models we develop for SLU are first pre-trained on task independent ASR data.  We use an ASR model trained on various telephone speech corpora \cite{george2021rnn}. The training process starts by training a CTC based acoustic model used to initialize the  transcription network of the RNN-T models. We use the recipe steps presented in \cite{audhkhasi2019forget,kurata2019guiding} for this model. The pre-trained RNN-T model we use for our experiments has a transcription network which contains 6 bidirectional LSTM layers with 640 cells per layer per direction. The prediction network is a single unidirectional LSTM layer with only 1024 cells. The joint network projects the 1280-dimensional stacked encoder vectors from the last layer of the transcription net and the 1024-dimensional prediction net embedding each to 256 dimensions, combines them multiplicatively, and applies a hyperbolic tangent. Finally, the output is projected to 42 logits, corresponding to 41 characters plus BLANK, followed by a softmax. More details on training settings and design choices can be found in \cite{george2021rnn}. The RNN-T based SLU models are trained using 40-dimensional,  global mean and variance normalized log-Mel filterbank features, extracted every 10 ms. These features are augmented with  $\Delta$ and $\Delta\Delta$ coefficients, every two consecutive frames are stacked, and every second frame is skipped, resulting in 240-dimensional vectors every 20 ms.

\begin{table}[t]
\caption{Evaluation results of BERT for the dialog-act task.}
\centering
  \resizebox{\columnwidth}{!}{
\begin{tabular}{cl|c}
\toprule
\multicolumn{2}{c}{Input Features} & F1 \\
\hline \hline
[B1] & Current-utterance & 56.9 \\ \hline
\hline
[B2] & + $[$SPEAKER$]$ + $[$HISTORY$]$ & 62.1\\ \hline
[B3] & + $[$SPEAKER$]$ + $[$DIALOG-ACTS$]$ & 60.5\\ \hline
[B4] & + $[$SPEAKER$]$ + $[$DIALOG-ACTS$]$ + $[$HISTORY$]$ & 64.4\\ \hline
[B5] & \MyIndentA - Last 4 utterances & 62.9\\ \hline
[B6] & \MyIndentA - Last 2 utterances & 61.9 \\ \hline

\end{tabular}}
\label{text-results-dialog-act-table-sam-dataset}
\vspace{-10pt}
\end{table}

\begin{table}[!hbt]
\caption{Performance on the dialog act task using the RNN-T SLU model with integrated dialog history.}

\centering
\begin{tabular}{cl|c}
\toprule
\multicolumn{2}{c}{Input Features} & F1 \\
\hline \hline
[C1] & Current-utterance (speech features) & 68.5\\ \hline \hline
\multicolumn{3}{l}{+ $[$SPEAKER$]$ + $[$DIALOG-ACTS$]$ BERT} \\ \hline
[C2]& Train on REF/Test on REF & 72.7 \\ \hline
[C3]& Train on REF/Test on DEC & 64.9 \\ \hline
[C4]& Train on DEC/Test on DEC & 65.5 \\ \hline \hline
\multicolumn{3}{l}{+ $[$SPEAKER$]$ + $[$HISTORY$]$ BERT} \\ \hline
[C5]& Train on REF/Test on REF & 71.3 \\ \hline
[C6]& Train on REF/Test on DEC & 66.8 \\ \hline
[C7]& Train on DEC/Test on DEC & 70.4 \\ \hline \hline
\multicolumn{3}{l}{+ $[$SPEAKER$]$ + $[$HISTORY$]$ + $[$DIALOG-ACTS$]$ BERT} \\ \hline
[C8]&Train on REF/Test on REF & 74.0 \\ \hline
[C9]&Train on REF/Test on DEC & 71.2 \\ \hline
[C10]& Train on DEC/Test on DEC & 72.1 \\ \hline
\end{tabular}
\label{speech-dc}
\vspace{-6pt}
\end{table}


For our experiments on the dialog act task, the pre-trained RNN-T model is first modified to accommodate new SLU labels for this task. As described earlier, during SLU adaptation, new network parameters are randomly initialized while the remaining parts of the network are copied from the pre-trained  network. For the dialog action prediction task, 16 additional nodes are added. The model is trained for 20 epochs on the augmented speech data set using an AdamW optimizer and a batch size of 16. The maximum learning rate is set to 2e-4 and a OneCycleLR policy which consists of a linear warm-up phase from 5e-5 to 2e-4 over the first 6 epochs followed by a linear annealing phase to 0 for the next 14 epochs, is employed \cite{smith2018superconvergence}. After this joint ASR+SLU training, the baseline model is used to decode the SLU test set. The model achieves a word error rate (WER) of 4.8\% along with an F1 score of 68.5\% for dialog act recognition. This result is reported as [C1] in Table \ref{speech-dc}.

We next conduct experiments using the three different kinds of dialog history embeddings discussed above. These 128 dimensional BERT embeddings are used as input features by appending them to the 240 dimensional acoustic features used to train the baseline system. Each of these three kinds of embedding are further tested in 3 different train/test settings. In the first setting, ground truth transcripts and dialog acts are used to create the embeddings (Train on REF/Test on REF). The performance results in this setting are an upper bound, as the embeddings are created from the best possible inputs. Because ground truth references are not available during test, we next use embeddings created using outputs from the baseline system in [C1] as dialog history (Train on REF/Test on DEC). In a third setting, we train a system with dialog history created using the baseline system in [C1] instead of ground truth references. This system is tested with decoded transcripts and dialog acts (Train on DEC/Test on DEC) and allows us to observe how much mismatch between train and test can be mitigated compared to the previous mismatched setting.

In experiments [C2], [C5] and [C8] the F1 scores improve significantly in all cases over the baseline results [C1], clearly demonstrating the benefit of integrating dialog history for this task. It is interesting to observe that history embeddings created using dialog acts [C2] perform better than embeddings created from transcripts [C5]. This is probably because dialog acts are a more compact representation of history than verbatim text.  In [C8], the results further improve when dialog acts are combined with transcript history to create embeddings. When these embeddings are tested in mismatched settings using decoded outputs, the performance however falls as seen in [C3] and [C6]. In contrast, the combination of dialog acts and transcripts in [C9] achieves better than individual results in mismatched settings. To alleviate the effect of train/test mismatch, we train models on decoded outputs in [C4], [C7] and [C10]. This helps improve performance as we obtain results closer to the oracle experiments tested in matched settings.

Similar to the dialog act task, for the caller intent task, the pre-trained RNN-T model is modified to accommodate new SLU labels for this task and then trained as above. In this case we add 8 extra output nodes to the pre-trained network as intent targets. As described in Section 3.2, a 128 dimensional history vector is created for each utterance using the ground truth transcripts or decoded outputs of all preceding utterances. Unlike the embedding vectors of the dialog act task, we do not use any additional tags other than text for creating these BERT embeddings. Similar to the experiments on the dialog act task, we evaluate the history embeddings in 3 train/test settings. A baseline SLU model constructed with just acoustic features corresponding to the current utterance being processed performs at 70.7\% accuracy. This result is reported as [D1] in Table \ref{speech-intent}.

\begin{table}[!hbt]
\caption{Performance on intent classification using the RNN-T SLU model with integrated dialog history.}

\centering
\begin{tabular}{cl|c}
\toprule
\multicolumn{2}{c}{Input Features} & Accuracy (\%) \\
\hline \hline
[D1] & Current-utterance (speech features) & 70.7\\ \hline \hline
\multicolumn{3}{l}{+ $[$HISTORY$]$ BERT} \\ \hline
[D2]&Train on REF/Test on REF & 94.1 \\ \hline
[D3]&Train on REF/Test on DEC & 93.9 \\ \hline
[D4]& Train on DEC/Test on DEC & 94.0 \\ \hline
\end{tabular}
\label{speech-intent}
\end{table}

In experiments [D2], [D3] and [D4] reported in Table \ref{speech-intent}, embedding vectors corresponding to each utterance's history are appended to the input acoustic features to train various SLU models. When ground truth transcripts are used to create history embeddings, we observe a very significant improvement in intent recognition performance, with the model performing at 94.0\% accuracy versus baseline [D1] 70.7\%, about 30\% relative improvement. Given that the WER of the baseline is quite low, the model performance does not degrade in [D3] when tested in mismatched settings with history vectors created using decoded transcripts at test time. Similar to the dialog act case, we train a separate system using history vectors created from decoded transcripts. This helps the model perform better in [D4].

In Tables 1-4 we have presented dialog act and intent recognition using both BERT and speech based SLU models. Although the results are on the same test sets, there is a significant gap between the BERT and speech based SLU results. 
From our analysis, we hypothesize that these differences are because the data set is inherently quite small, the turns are scripted, there is only a limited set of speakers, and there are several inconsistencies in the annotations. The speech model has to a good extent been able to memorize many details of speakers and characteristics of the dataset and hence these differences. Results within each training methodology are however consistent and we see  clear benefits from our proposed approaches. 

\section{Conclusions}
 In this paper we have presented a comprehensive study on the use of dialog history for E2E SLU systems. We investigated the use of several kinds of BERT based embeddings to capture important information available in an utterance's dialog history. These embeddings were then used as input features to significantly improve the performance of SLU systems on two tasks. We have shown that although SLU labels like dialog acts can be as effective as verbatim transcripts, embeddings created with just these labels might not be robust to recognition errors. In future work we plan to investigate the effect of noisy transcripts in creating robust dialog history embeddings. Our proposed approaches improve SLU performance by up to 8-30\% on the two tasks we have investigated.

\bibliographystyle{IEEEtran}

\bibliography{mybib}

\end{document}